\documentclass{article}

\usepackage{arxiv}
\usepackage{algorithm}
\usepackage{algpseudocode}
\usepackage{amssymb}
\usepackage{amsmath}
\usepackage{graphicx}
\usepackage{subfig}
\usepackage{pgfplots}
\usepackage{soul}
\usepackage{cite}
\pgfplotsset{compat=1.18}

\title{Sliding Window Sum Algorithms for Deep Neural Networks}
\author{
 Roman Snytsar \\
 AI \& Research \\
 Microsoft \\
 Redmond WA 98052 \\
 \texttt{Roman.Snytsar@microsoft.com}
}

\date{May 2023}

\begin{document}

\maketitle

\begin{abstract}
{Sliding window sums are widely used for string indexing, hashing and time series analysis. We have developed a family of the generic vectorized sliding sum algorithms that provide speedup of $O(P/w)$ for window size $w$ and number of processors $P$. For a sum with a commutative operator the speedup is improved to $O(P/log(w))$. Even more important, our algorithms exhibit efficient memory access patterns.\\
In this paper we study the application of the sliding sum algorithms to the training and inference of the Deep Neural Networks. We demonstrate how both pooling and convolution primitives could be expressed as sliding sums and evaluated by the compute kernels with the shared structure.\\
We show that the sliding sum convolution kernels are more efficient than the commonly used GEMM kernels on the CPU, and could even outperform their GPU counterparts.
}

\end{abstract}
\vspace{0.35cm}

\section{Introduction}

A Deep Neural Network (DNN) is one of the most significant tools in the arsenal of the machine learning (ML) researcher\cite{li2021survey}. DNNs are constructed from multiple layers that transform the data sequentially via operations such as pooling, convolution, and activation. In most successful DNNs the great portion of computational resources is consumed performing convolution.\\
A common approach to implementing convolutional layers is to expand the input into a column matrix (\emph{im2col}) and then call a highly tuned General Matrix Multiplication (GEMM) procedure from the existing linear algebra library such as BLIS\cite{van2015blis} or MKL\cite{wang2014intel}. Since the hardware optimized GEMM implementations exist for every standard CPU, graphics processing unit (GPU), or digital signal processor (DSP), the \emph{im2col} approach has been highly successful in DNN frameworks such as Caffe\cite{jia2014caffe}, ONNX\cite{onnx2023} and Torch\cite{collobert2002torch}. The popularity of the \emph{im2col} tactics also influences the design of the custom hardware marketed as ML accelerators that are in fact the GEMM accelerators.\\
The major disadvantages of the \emph{im2col} conversion are the greatly increased memory footprint of the input matrix and the reduced data locality. For a convolution with a filter size k, the column matrix is k times larger than the original input. A lot of effort has been put into remediating this problem. A mainstream approach is converting input to the low precision floating point or even integer representation\cite{gholami2021survey}. The quantization techniques reduce the memory footprint and latency by an order of magnitude and even influence the design of the GEMM accelerators. It is important to note though that the quantization is not entangled with GEMM and could be equally successfully applied to the original convolution problem. Yet another research trend is applying the GEMM routines to the smaller intermediate data structures\cite{anderson2017low} \cite{vasudevan2017parallel}or even to the original input data\cite{wang2021optimization}.\\
We propose an approach that replaces GEMM with a new kind of the computation kernel that operates on the unmodified input. 

\section{Methods}
\label{sec:method}
\subsection{Prefix Sum}

At the foundation of our method is the concept of a \emph{prefix sum}, and the accompanying \emph{reduce} and \emph{scan} algorithms.
A prefix sum is a transformation that takes an operator $\oplus$, and a sequence of elements
\[ x_0, x_1, \ldots, x_k, \ldots \]
and returns the sequence
\begin{equation}\label{prefixSum}
y_i=\sum\limits_{j=0}^{i} x_j = x_0\oplus x_1\oplus\ldots\oplus x_i
\end{equation}
or in recurrent form 
\begin{equation}\label{prefixSumRecur}
y_{i+1}=y_i\oplus x_{i+1}
\end{equation}
Despite the data carry dependency, the $Nth$ element of the prefix sum with an associative operator could be computed in $O(log(N))$ parallel steps using \emph{reduce} algorithm. Even stronger statement is that all $N$ elements of the prefix sum could be computed in the same $O(log(N))$ parallel steps using \emph{scan} algorithm, as shown by \cite{Blelloch93}. 

\subsection{Sliding Window Sum}

Sliding window sum (sliding sum) takes a window size $w$ in addition to an operator $\oplus$, and a sequence of elements, and returns the sequence
\begin{equation}\label{slidingSum}
y_i=\sum\limits_{j=i}^{i+w-1} x_j = x_i\oplus x_{i+1}\oplus\ldots\oplus x_{i+w-1}
\end{equation}
where each sum is defined in terms of the operator $\oplus$ and contains exactly $w$ addends.
The asymptotic complexity of a naive sliding sum algorithm is $O(wN)$ where $N$ is the length of the source sequence.

Every sum defined by Equation~\ref{slidingSum} is a prefix sum with operator $\oplus$ and input sequence $x_i\ldots\oplus x_{i+w-1}$. Many useful operators are associative, so the prefix scan algorithm is applicable here, reducing complexity of every sum in Equation~\ref{slidingSum} to $O(log(w))$ and, trivially, the overall sliding sum complexity to $O(Nlog(w))$ parallel steps.

While working on the bioinformatics applications, we have used sliding window sums to represent the minimizer seeds and have developed a family of algorithms to achieve better parallel speedups\cite{snytsar2020parallel}. Now we will apply the same approach to the DNN operators.

\subsection{Pooling}

The average pooling operator is trivially the sliding window sum with the associative operator $+$. By analogy, the max pooling operator is a sliding window sum with the associative operator $max$. Implementing the sliding pooling operator could be a warm-up before concentrating on the convolution.

\subsection{Dot Product}
As a corollary we will show that a dot product is a prefix sum. Dot product of the two vectors of length $M$ is defined as:
\begin{equation}\label{dotProduct}
c=\sum\limits_{i=0}^{M-1} a_ib_i
\end{equation}
First, we replace vectors $a$ and $b$ with vectors $\alpha$ and $\beta$ so that:
\begin{equation}\label{dotMasking}
\alpha_i=\begin{cases}1, & a_i=0\\a_i, & otherwise\end{cases},  \beta_i=\begin{cases}0, & a_i=0\\b_i, & otherwise\end{cases}
\end{equation}
It holds that 
\begin{equation}\label{dotIdentity}
\sum_{i=0}^{M-1}{\alpha_i\beta_i}=\sum_{i=0}^{M-1}{a_ib_i}
\end{equation}
Next, we define a sequence of $(M+1)$ pairs,
\begin{equation}\label{gammaSequence}
\gamma_i=\begin{pmatrix}u_i\\v_i\end{pmatrix},\quad where \quad u_i=\begin{cases}
  1, & i=0 \\  \frac{\alpha_{i-1}}{\alpha_i}, & 0<i<M\\ \alpha_{M-1}, & i=M
\end{cases},\quad and \quad v_i=\begin{cases}\beta_i, & i<M\\0, & i=M
\end{cases}
\end{equation}
and the operator $\oplus$ such that
\begin{equation}\label{gammaOperator}
\gamma_i\oplus\gamma_j=\begin{pmatrix}u_i\\v_i\end{pmatrix}\oplus\begin{pmatrix}u_j\\v_j\end{pmatrix}=\begin{pmatrix}u_i\bullet u_j\\u_j\bullet v_i+v_j\end{pmatrix}
\end{equation}
Operator $\oplus$ is associative \cite{Blelloch93}, and the sequence 
\begin{equation}\label{gammaSum}
\delta_i=\begin{cases}\gamma_0, & i=0\\\delta_{i-1}\oplus\gamma_i, & 0<i\leq M\end{cases}
\end{equation}
is a prefix sum. The bottom element of the last sum $\delta_M$ is the original dot product, and $\delta_M$ could be evaluated using reduce algorithm in $log(M)$ parallel steps of fused multiply-add 
(FMA) operations. The total work is still $M$. The important result here is representing dot product as a prefix sum in preparations for the sliding window implementation of the convolution.

\subsection{Convolution}
Convolution is a sliding window sum (dot product) with the associative operator defined by equation \ref{gammaOperator}. Consequently our family of the sliding window algorithms is applicable to the convolution operator.

\section{Algorithms}
\label{sec:algorithm}
\subsection{Vector Algorithms}

Our first algorithm is a vector-friendly way of calculating sliding sum assuming the input sequence elements become available one by one and are processed using the vector instructions of width $P>w$:
\begin{algorithm}
	\caption{Scalar Input}\label{al:scalar}
	\begin{algorithmic}[0]
		\Procedure{ScalarInput}{$x_0\dots x_{n-1}$}
		\State $Y\gets \Big(\underbrace{\sum\limits_{j=0}^{w-2} x_j, \sum\limits_{j=1}^{w-2} x_j, \dots , x_{w-3}\oplus x_{w-2}, x_{w-2}}_{w-1}, 0, \dots , 0\Big)$
		\For{$i=w-1$ to N}
		\State $X\gets \Big(\underbrace{x_i, x_i, \dots , x_i}_w, 0, \dots , 0\Big)$
		\State $Y\gets Y \oplus X$
		\State $y_{i-w+1}\gets Y[0]$
		\State $Y\gets Y \lll 1$
		\EndFor
		\EndProcedure
	\end{algorithmic}
\end{algorithm}

Vector Y is initialized to the suffix sums with the number of elements decreasing from $w-1$ to $0$. Then in a loop every incoming element $x_k$ is broadcast to the first $w$ elements of vector X. After vector addition the zeroth element of Y contains the next sliding sum. Next, the vector Y is shifted left by one element, as denoted by operator $\lll$, and the state is ready for the next iteration.

Asymptotic complexity of the scalar input algorithm is $O(N)$ with no additional requirements on the operator $\oplus$.

This result could be improved if we assume that the input sequence arrives packed in vectors of width $P>w$.
\begin{algorithm}
\caption{Vector Input}\label{al:vector}
\begin{algorithmic}[0]
\Procedure{VectorInput}{$x_0\dots x_{N-1}$}
\State $Y\gets \Big(\underbrace{\sum\limits_{j=0}^{w-2} x_j, \sum\limits_{j=1}^{w-2} x_j, \dots , x_{w-3}\oplus x_{w-2}, x_{w-2}}_{w-1}, 0, \dots , 0\Big)$
\For{$i=w-1$ to N step P}
\State $X\gets \Big(x_i, x_{i+1}, \dots , x_{i+p-1}\Big)$
\State $X1\gets \Big(\underbrace{X[0], X[0]\oplus X[1], \dots , \sum\limits_{j=0}^{w-2} X[j]}_{w-1}, \sum\limits_{j=0}^{w-1} X[j], \dots , \sum\limits_{j=p-w}^{p-1} X[j]\Big)$
\State $Y1\gets \Big(0, \dots , 0, \underbrace{\sum\limits_{j=p-w}^{p-1} X[j], \sum\limits_{j=p-w}^{p-2} X[j], \dots , X[p-w]}_{w-1} \Big)$
\State $Y\gets Y \oplus X1$
\State $ y_{i-w+1}\dots y_{i-w+p}\gets Y[0]\dots Y[p-1]$
\State $Y\gets Y1 \lll (P-w)$
\EndFor
\EndProcedure
\end{algorithmic}
\end{algorithm}

At every iteration $P$ input elements are placed into vector $X$. X1 is filled with the prefix sums of up to $w$ addends, and Y1 is filled with the suffix sums constructed from the elements of $X$. Then the vector sum of $Y$ and $X1$ yields the next P output elements. Finally, the suffix sums from $Y1$ are shifted into proper positions in vector $Y$, and it is ready for the next iteration.

The asymptotic complexity thus is  $O(N\cdot w/P)$ with the parallel speedup $O(P/w)$ for any operator $\oplus$. If $\oplus$ is associative, the prefix/suffix sums could be computed in parallel using the algorithm in \cite{Blelloch93}, and the complexity is reduced to $O(N\cdot log(w)/P)$ with the speedup improving to $O(P/log(w))$.

For example, since $min$ is an associative operator, the sliding window minimum can be computed using the faster version of the vector input algorithm.

One might notice that the suffix sum computation utilises only $w-1$ out of $p$ elements of vector $Y1$.  Eliminating this inefficiency leads to the Ping Pong algorithm where both suffix and prefix sums yield some elements of the output sequence.

\begin{algorithm}
\caption{Ping Pong}\label{al:pingpong}
\begin{algorithmic}[0]
\Procedure{PingPong}{$x_0\dots x_{N-1}$}
\For{$i=0$ to N step 2P-w+1}
\State $Y\gets \Big(x_i, x_{i+1}, \dots , x_{i+p-1}\Big)$
\State $X\gets \Big(x_{i+P}, x_{i+P+1}, \dots , x_{i+2P-1}\Big)$
\State $Y1\gets \Big(\underbrace{\sum\limits_{j=0}^{w-1} Y[j], \sum\limits_{j=1}^{w} Y[j], \dots , \sum\limits_{j=P-w}^{P-1} Y[j]}_{P-w+1}, \sum\limits_{j=P-w}^{P-1} Y[j], \dots , Y_{P-1} \Big)$
\State $ y_i\dots y_{i +P-w}\gets Y[0]\dots Y[P-w]$
\State $Y1\gets Y1 \lll (P-w)$
\State $X1\gets \Big(\underbrace{X[0], X[0]\oplus X[1], \dots , \sum\limits_{j=0}^{w-2} X[j]}_{w-1}, \sum\limits_{j=0}^{w-1} X[j], \dots , \sum\limits_{j=p-w}^{p-1} X[j]\Big)$
\State $Y1\gets Y1 \oplus X1$
\State $ y_{i+P-w+1}\dots y_{i+2P - w}\gets Y[0]\dots Y[P-1]$
\EndFor
\EndProcedure
\end{algorithmic}
\end{algorithm}

The Ping Pong algorithm does not offer any asymptotic improvements but is 30-50\% faster in practice. It however accesses memory in strides unaligned to $P$, and the two memory loads per iteration present a challenge while implementing boundary conditions like padding, mirroring, or periodicity.\\
A simpler algorithm could be formulated in terms of the Slide operation that extracts a vector from the concatenation of the two inputs. It directly maps to the {\bf EXT} instruction from the ARM SVE instruction set \cite{stephens2017arm}, and is easily implemented via appropriately named {\bf vslideup/down} instructions of the RISC-V vector extension \cite{rvv2023}, or the {\bf vperm*2ps} Intel AVX512 instructions \cite{intel2023}. Similar to the previous algorithms, if $\oplus$ is associative, the inner loop could be replaced by the parallel reduction for maximum parallel speedup.
\begin{algorithm}
\caption{Vector Slide}\label{al:vectorslide}
\begin{algorithmic}[0]
\Procedure{Slide}{$Y1$, $Y2$, $offset$}
\State $Y\gets CONCAT(Y1, Y2)$
\State \Return $\big(\underbrace{Y[offset], Y[offset+1], \dots, Y[offset+P-1]}_P\big)$
\EndProcedure
\Procedure{VectorSlide}{$x_0\dots x_{N-1}$}
\State $Y\gets (x_0, x_1, \dots, x_{P-1})$
\For{$i=P$ to $N$ step $P$}
\State $Y1\gets \Big(x_i, x_{i+1}, \dots , x_{i+p-1}\Big)$
\State $X\gets Y1$
\For{$k=1$ to $w-1$}
\State $X\gets X \oplus $Slide$(Y, Y1, P-k)$
\EndFor
\State $ y_{i}\dots y_{i+p-1}\gets X[0]\dots X[p-1]$
\State $Y\gets Y1$
\EndFor
\EndProcedure
\end{algorithmic}
\end{algorithm}

\section{Experiments}
\begin{figure}[!tpb]
	\centering
	\begin{tikzpicture}
		\begin{axis}[
            width=.8\textwidth,
			xlabel={Filter Size}]
			\addplot[mark=square*] coordinates {
				(3,1.78)
				(5,2.18)
				(11,3.6)
				(17,4.75)
				(21,5.71)
				(29,7.17)
				(37,8.12)
				(49,8.52)
				(51,8.37)
			};
		\end{axis}
	\end{tikzpicture}
	\caption{Speedup of the 1-D Convolution.}\label{speedup1d}
\end{figure}
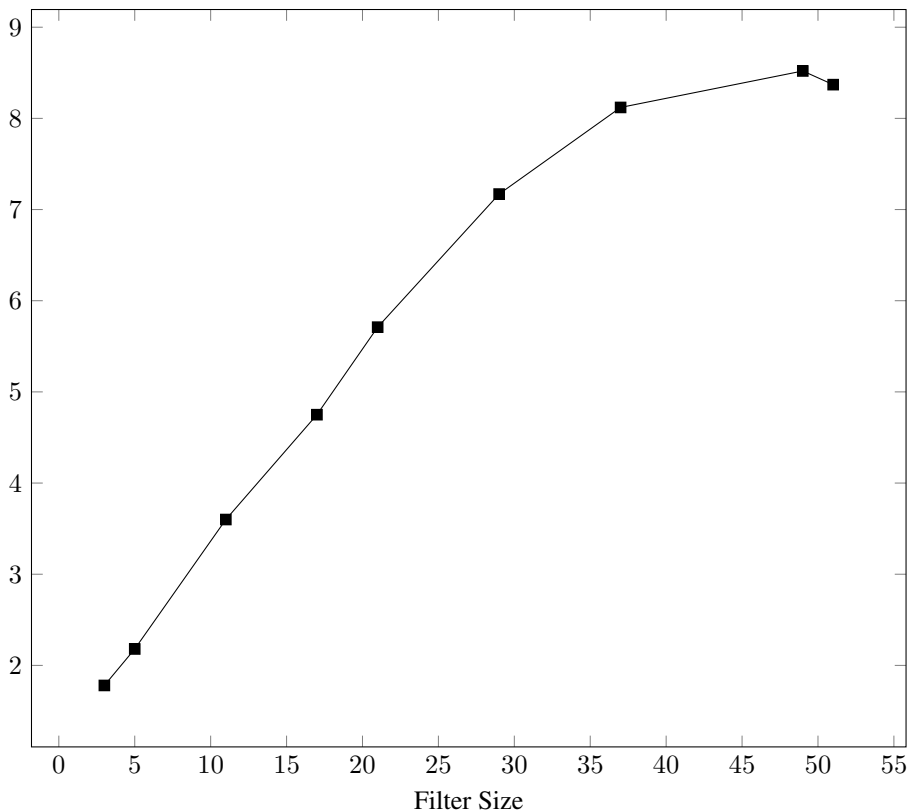

We have implemented the sliding window convolution as an alternate execution path in the ONNX framework \cite{onnx2023}. Figure \ref{speedup1d} shows the achieved speedup when compared to the baseline |emph{MlasConv} procedure applied to the large 1-D input and the filters of various sizes. The speedup is approximately proportional to the logarithm of the kernel size. \\
Additionally, we have recreated the scenario for the very large dilated kernel described in \cite{chaudhary2021efficient}. We have achieved up to 6.8x speedup on the small data set and around 4x speedup across the board approaching the GPU performance numbers.\\
\begin{figure}[!tpb]
	\centering
	\begin{tikzpicture}
		\begin{axis}[
            width=.8\textwidth,
            xmode=log,
			xlabel={Output Size}]
			\addplot[mark=square*] coordinates {
				(1000,6.845946984)
				(5000,4.399866427)
				(10000,4.312676076)
				(50000,4.013980206)
				(100000,3.393491474)
			};
		\end{axis}
	\end{tikzpicture}
	\caption{Dilated Convolution Speedup.}\label{dilated1d}
\end{figure}
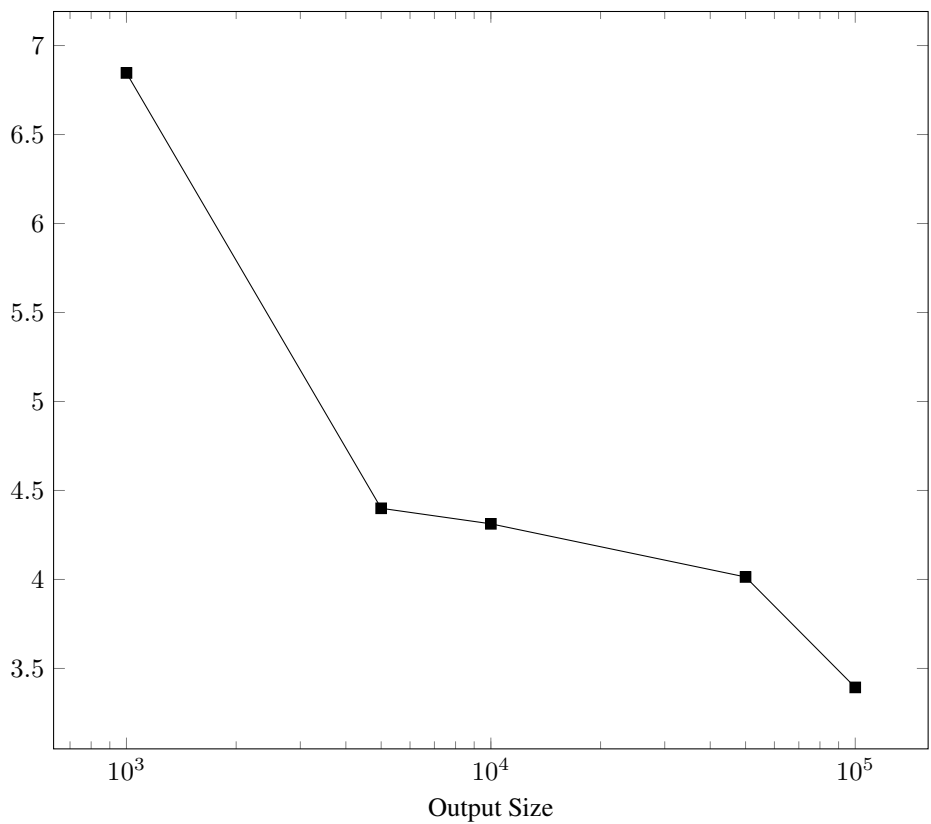
\section{Conclusion}
We have demonstrated excellent performance of the Sliding Sum algorithms using the commodity hardware without the need for the specialized accelerators. Despite the promising results, there is plenty of work ahead.\\
The most obvious next step is extending the sliding convolution approach to more than one dimension covering the majority of the DNN applications.\\
The most common filter sizes in the DNN applications are 3 an 5 in every dimension. With the filter this small the current sliding convolution algorithms demonstrate very modest speedup since the number of the arithmetic instructions per the memory load is low. The situation improves in the multiple dimensions but still could require custom compute kernels for the small filter sizes.\\
Lastly, since the  accelerators for the matrix multiplication are already present in the current generation of the hardware, it would be wise re-using them. Thus, it is important to re-formulate our algorithms in terms of the small matrix multiplication completing the circle.\\
\bibliographystyle{plain}
\bibliography{ConvolutionSlidingSum}
\end{document}